From Complexity to Clarity:
How AI Enhances Perceptions of Scientists and the Public's Understanding of Science

David M. Markowitz[1]

**Affiliations**
[1] Department of Communication, Michigan State University, East Lansing, MI 48824 USA

Email: dmm@msu.edu

**This paper is currently *in press* at PNAS Nexus. Please use the following citation for this paper:**

**Markowitz, D. M. (in press). From complexity to clarity: How AI enhances perceptions of scientists and the public's understanding of science. *PNAS Nexus*.**

**Abstract**
This paper evaluated the effectiveness of using generative AI to simplify science communication and enhance the public's understanding of science. By comparing lay summaries of journal articles from PNAS, yoked to those generated by AI, this work first assessed linguistic simplicity differences across such summaries and public perceptions in follow-up experiments. Specifically, Study 1a analyzed simplicity features of PNAS abstracts (scientific summaries) and significance statements (lay summaries), observing that lay summaries were indeed linguistically simpler, but effect size differences were small. Study 1b used a large language model, GPT-4, to create significance statements based on paper abstracts and this more than doubled the average effect size without fine-tuning. Study 2 experimentally demonstrated that simply-written GPT summaries facilitated more favorable perceptions of scientists (they were perceived as more credible and trustworthy, but less intelligent) than more complexly-written human PNAS summaries. Crucially, Study 3 experimentally demonstrated that participants comprehended scientific writing better after reading simple GPT summaries compared to complex PNAS summaries. In their own words, participants also summarized scientific papers in a more detailed and concrete manner after reading GPT summaries compared to PNAS summaries of the same article. AI has the potential to engage scientific communities and the public via a simple language heuristic, advocating for its integration into scientific dissemination for a more informed society.

**Significance Statement**
Across several studies, this paper revealed that generative AI can simplify science communication, making complex concepts feel more accessible and enhancing public perceptions of scientists. By comparing traditional scientific summaries from the journal PNAS to AI-generated summaries of the same work, this research demonstrated that AI can produce even simpler and clearer explanations of scientific information that are easier for the general public to understand. Importantly, these simplified summaries can improve perceptions of scientists and their understanding of the science as experimentally demonstrated in this work. With small, language-level changes, AI can facilitate effective science communication and its possible deployment at scale makes it an appealing technology for authors and journals.

# From Complexity to Clarity: How AI Enhances Perceptions of Scientists and the Public's Understanding of Science

Scientific information is essential for everyday decision-making. People often use science, or information communicated by scientists, to make decisions in medical settings (1), environmental settings (2), and many others (3). For people to use such information effectively, however, they must have some amount of scientific literacy (4) or at least trust those who communicate scientific information to them (5). Overwhelming evidence suggests these ideals are not being met, as trust in scientists and scientific evidence have decreased over time for nontrivial reasons (e.g., distrust in institutions, political polarization, among many others) (6–8). The public's decreasing trust in scientists and scientific information is unrelenting, which requires more thoughtful research into countermeasures and possible remedies that can be scaled across people and populations.

Several remedies have been proposed to make science more approachable, and to improve the perception of scientists. For example, some propose that being transparent about how research was conducted and disclosing possible conflicts of interest (9, 10), having scientists engage with the public about their work (11), or improving scientists' ability to tell a compelling story (12) can increase public trust. While there is no panacea for dwindling public perceptions of science and scientists, extant evidence suggests this is an issue worth taking seriously, and it is imperative that scientists discover ways to best communicate their work with the hope of improving how people consider them and their research.

Against this backdrop, the current work argues that how one's science is communicated matters, and that language-level changes to scientific summaries that make the writing simpler (e.g., using more common words with shorter sentences) can significantly improve perceptions of a scientist. Critically, the evidence suggests scientists may not be the best messengers to communicate their work if one goal is to communicate science simply. In other words, it may be difficult for experts to write for non-experts. Instead, as the current research demonstrates, generative AI can effectively summarize scientific writing in ways that are more approachable for lay readers, and such tools can be scaled to improve science communication at a system level and comprehension at a person level.

## The Benefits of Simple Writing

The idea that simple language patterns can improve perceptions of scientists is supported by decades of processing fluency research and feelings-as-information theory (13–16). This literature suggests people tend to use their feelings when consuming information (16, 17), and people often prefer simplicity over complexity because simple (fluent) information feels better to most people than complex (disfluent) information. Support for this contention in the laboratory and the field indeed suggests people engage with, approach, and prefer content that is written in simple versus complex terms (e.g., simple synonyms of the same concept compared to complex synonyms) (18–22). Together, much of this research supports the *simpler-is-better* hypothesis (18) and a *simple writing heuristic* (21): people will engage with, prefer, and psychologically attend more to language that is communicated simply and fluently, absent some instrumental goal being activated (18, 23, 24).

The most common linguistic fluency dimension evaluated in the literature is lexical fluency, which considers the degree to which people use common and everyday terms in communication. People perceive scientists to be more intelligent if their work is written with simple words (e.g., the word *job*) compared to complex words (e.g., the word *occupation*) (15). In most cases, people prefer simple synonyms for a concept compared to complex synonyms of the same concept because it is more of a challenge to interpret and comprehend complexity, and people are economical with their effort and attention (22, 25). Another fluency dimension is analytic writing fluency (24). This dimension considers one's communication style and *how* people communicate, instead of what they are communicating about (26, 27). According to prior work, a simple communication style is informal and reflects a story (e.g., it contains more pronouns and adverbs) compared to a complex communication style, which is formal and contains high rates of articles and prepositions (28–30). Finally, another relevant fluency concept is structural fluency, which considers the length of words and sentences. Longer words (e.g., *occupation* vs. *job*) and sentences with more words tend to require more effort to process (31). The final marker of fluency relevant to the current work is operationalized by readability, which considers verbal simplicity/complexity in terms of word and sentence length.

**The Current Work**

The current empirical package evaluates fluency effects in the context of science writing and has several aims. The first aim is to evaluate if lay summaries of scientific articles (called significance statements in many journals) are indeed linguistically simpler than scientific summaries of the same articles (abstracts). It is unclear if scientists are aware of how to effectively summarize their work for non-experts (32), making it important to empirically test if ideals of a journal, like simple and approachable writing, are being realized (Study 1a). The second aim is to evaluate if such lay summaries can be made even simpler with generative Artificial Intelligence (AI), which has demonstrated immense potential to summarize and simplify science writing in prior work (33, 34). To this end, Study 1b had a popular large language model create lay summaries of a paper, comparing its linguistic properties to human counterparts, to identify how AI can facilitate more approachable and more simply-written science.

Finally, building on this progression of studies, two experiments tested the causal impact of reading scientific writing generated by AI, versus reading scientific writing by humans, on perceptions of scientists (Study 2) and participants' understanding of the science (Study 3). In Study 2, participants were randomly assigned to AI or human versions of a scientific summary, and they made judgments about the credibility, trustworthiness, and intelligence of the authors; in Study 3, they also summarized the science in their own words and answered a multiple-choice question about the research. To foreshadow the results: people preferred the simple (AI) versions compared to the complex (human) versions, yet ironically, people believed that the complex versions were more likely to be AI than human. Participants also comprehended the science better after reading simple, AI-generated scientific summaries compared to human-generated scientific summaries.

## Study 1a: Method

**Data Collection**

To first evaluate if lay summaries had a simpler linguistic style than scientific summaries, significance statements and academic abstracts were respectively extracted from the journal *Proceedings of the National Academy of Sciences* (PNAS). This journal was selected because it is a widely-read, high-impact general science journal that was one of the first outlets to require authors to provide traditional scientific summaries (e.g., abstracts) and lay summaries that appeal to average readers. PNAS also has topical breadth, scale, and longevity relative to other journals that may require lay summaries in that significance statements began in 2012 (35).

A total of 42,022 publications were extracted from PNAS between January 2010 and March 2024 to capture possible papers that included both academic abstracts and significance statements. Only those with both summary types were included in this paper to create a yoked comparison within the same article. The final dataset included 34,584 papers (34,584 significance statements and 34,584 abstracts), totaling 10,799,256 words.

**Automated Text Analysis**

All texts were evaluated with Linguistic Inquiry and Word Count (LIWC), an automated text analysis tool that counts words as a percentage of the total word count per text (36). LIWC contains a validated internal dictionary of social (e.g., words related to family), psychological (e.g., words related to cognition, emotion), and part of speech dimensions (e.g., pronouns, articles, prepositions), and the tool measures the degree to which each text contains words from its respective dictionary categories. For example, the phrase "This science aims to improve society" contains 6 words and counts the following LIWC categories, including but not limited to: impersonal pronouns (*this*; 16.67% of the total word count) and positive tone words (*improve*; 16.67% of the total word count). All texts were run through LIWC-22 unless otherwise stated.

**Measures**

To evaluate how lay versus scientific summaries compared in terms of verbal simplicity, three measures were used from prior work to approximate simple language patterns (24): common words (e.g., the degree to which people use common and simple terms like *job* instead of uncommon and more complex terms like *occupation*), one's analytic writing style (e.g., the degree to which people have a formal and complex writing style compared to an informal and narrative-like writing style), and readability (e.g., the number of words per sentence and big words in a person's communication output).

Consistent with prior work (24, 37–39), common words were operationalized with the LIWC dictionary category. LIWC's dictionary represents a collection everyday words in English (40, 41). Therefore, texts that use more words from this dictionary are simpler than texts that use fewer words from this dictionary. One's analytic writing style was operationalized with the LIWC analytic index, which is a composite variable of seven style word categories. Style words represent *how* one is communicating rather than what they are communicating about (26, 42). This index contains high rates of articles and prepositions, but low rates of conjunctions, adverbs, auxiliary verbs, negations, and pronouns (28, 43, 44).[1] Low scores (e.g., those that are less analytic) are stylistically simpler than high scores. Finally, readability was operationalized with the Flesch Reading Ease metric (31) and calculated using the *quanteda.textstats* package in R (45). High scores on the Flesch Reading Ease metric suggest more readable and simpler writing (e.g., texts with smaller words and shorter sentences) compared to low scores. These language dimensions were evaluated as an index by first standardizing (z-scoring) each variable and then applying the following formula: Common Words + Readability – Analytic Writing. High scores are linguistically simpler than low scores.

**Analytic Plan**

Since each article contained one lay summary and one scientific summary from the same article, independent samples t-tests were conducted for the simplicity index and each individual dimension of the index. All data across studies are located on the Open Science Framework (OSF): https://osf.io/64am3/?view_only=883926733e6e494fa2f2011334b24796. Descriptive statistics and intercorrelations for key variables are in the online supplement.

**Study 1a: Results**

As expected, lay summaries were linguistically simpler than scientific summaries of the same article, Welch's $t(65793) = 40.62$, $p < .001$, Cohen's $d = 0.31$, 95% CI [0.29, 0.32].[2] At the item level of the simplicity index, lay summaries ($M = 69.77\%$, $SD = 7.14\%$) contained more common words than scientific summaries ($M = 67.79\%$, $SD = 6.60\%$), Welch's $t(68741) = 37.79$, $p < .001$, Cohen's $d = 0.29$, 95% CI [0.27, 0.30]. Lay summaries ($M = 92.34$, $SD = 7.95$) also had a simpler linguistic style than scientific summaries ($M = 94.31$, $SD = 5.19$), Welch's $t(59561) = -38.52$, $p < .001$, Cohen's $d = 0.29$, 95% CI [0.28, 0.31]. Finally, lay summaries ($M = 12.96$, $SD = 13.93$) were more readable than scientific summaries as well ($M = 12.49$, $SD = 12.46$), Welch's $t(68320) = 4.67$, $p < .001$, Cohen's $d = 0.036$, 95% CI [0.02, 0.05].

[1] Analytic writing = [articles + prepositions - pronouns - auxiliary verbs - adverb - conjunctions - negations] from LIWC scores (44).

[2] 95% Confidence Intervals were bootstrapped with 5,000 replicates.

Together, while lay summaries were indeed linguistically simpler than scientific summaries at PNAS, the effect sizes between such groups were quite small and it is therefore unclear if individual readers would be able to recognize or appreciate such differences. Can lay summaries be written even simpler, using generative AI tools, to produce more substantive effect sizes while maintaining the core content of each text? In the next study, a random selection of abstracts was submitted to a popular large language model, GPT-4, and were given the same instructions as PNAS authors on how to construct a significance statement.

## Study 1b: Method

An *a priori* power analysis using a small effect size (Cohen's $d$ = 0.20) powered at 80% suggested 788 cases were needed to detect a difference between GPT significance statements and PNAS significance statements. This study was oversampled, and a random selection of 800 abstracts from Study 1a was used in this study to create a comparison of 800 PNAS significance statements to 800 AI-generated significance statements based on PNAS abstracts ($N$ = 1,600 total texts). Using the OpenAI API, the large language model GPT-4 was fed each abstract individually and given the following prompt, which was drawn from descriptions of what PNAS authors should communicate in their significance statements (35):

> *The following text is an academic abstract from the journal Proceedings of the National Academy of Sciences. Based on this abstract, create a significance statement. This statement should provide enough context for the paper's implications to be clear to readers. The statement should not contain references and should avoid numbers, measurements, and acronyms unless necessary. It should explain the significance of the research at a level understandable to an undergraduate-educated scientist outside their field of specialty. Finally, it should include no more than 120 words. Write the significance statement here*:

The same text analytic process was performed on these data as Study 1a. Each GPT significance statement then received scores based on common words (LIWC dictionary), analytic writing (LIWC analytic writing), and readability (Flesch Reading Ease).

## Study 1b: Results

Distributions of the comparisons in this study are reflected in Figure 1. Indeed, GPT significance statements were written in a simpler manner than PNAS significance statements for the simplicity index, Welch's $t(1492.1) = 11.55$, $p < .001$, Cohen's $d = 0.58$, 95% CI [0.47, 0.69]. Specifically, GPT significance statements ($M$ = 75.53%, $SD$ = 5.57%) contained more common words than PNAS significance statements ($M$ = 69.84%, $SD$ = 7.45%), Welch's $t(1478.7) = 17.31$, $p < .001$, Cohen's $d = 0.87$, 95% CI [0.76, 0.97]. GPT significance statements ($M$ = 17.59, $SD$ = 11.15) were also more readable than PNAS significance statements ($M$ = 12.86, $SD$ = 14.27), Welch's $t(1510) = 7.39$, $p < .001$, Cohen's $d = 0.37$, 95% CI [0.27, 0.47]. However, GPT significance statements ($M$ = 92.73, $SD$ = 6.89) had a statistically equivalent analytic style as PNAS significance statements ($M$ = 92.32, $SD$ = 7.48), Welch's $t(1587.7) = 1.16$, $p = .246$, Cohen's $d = 0.06$, 95% CI [-0.04, 0.16].

### Alternative Explanations

One possible explanation for the Study 1b results is that there were content differences across the PNAS and GPT texts explaining or impacting such simplicity effects. This concern was addressed in two ways. First, PNAS has various sections that authors submit to, and LIWC has categories to approximate words associated with such sections. For example, the LIWC category for political speech would approximate papers submitted the Social Science section, specifically Political Sciences. Several linguistic covariates were therefore examined to account for content-related differences across GPT and PNAS texts. After including overall affect/emotion and cognition (to control for topics within the Psychological Sciences section of PNAS), political speech (to control for topics within the Political Science section of PNAS), and physical references to the multivariate models (to control for topics within the Biological Sciences section of PNAS), all results were maintained except for Analytic writing, where GPT texts were more analytic than PNAS texts, which is also consistent with prior work (46). Please see the online supplement for additional LIWC differences across these text types.

Content effects were also evaluated in a bottom-up manner using the Meaning Extraction Method to measure dominant themes across the GPT and PNAS texts (47, 48). The evidence in the online supplement states there were 8 themes reliably extracted from the data, ranging from basic methodological and research information to gene expression and cancer research. Controlling for these themes, including the prior LIWC content dimensions, revealed consistent results as well (see supplement). Therefore, Study 1b evidence is robust to content.

Altogether, human authors write simpler for lay audiences than for scientific audiences (Study 1a), but Study 1b demonstrated artificial intelligence and large language models can do so more effectively (e.g., the effect size differences between GPT significance statements and PNAS significance statements were larger than that of humans in Study 1a). The findings thus far are correlational and therefore need causal evidence to demonstrate the impact of these effects on human perceptions and comprehension. In Study 2, participants were randomly assigned to read a GPT significance statement or PNAS significance statement from pairs of texts that appeared in the previous studies. Participants made perceptions about the author (e.g., intelligence, credibility, trustworthiness), judged the complexity of each text, and they rated how much they believed the author of each text was human or artificial intelligence. Only perceptions of the author were made because prior work suggests people generally report consistent ratings when asked about both scientists and their science in similar studies (9).

## Study 2: Method

Participants in the US were recruited from Prolific and paid $4.00 for their time in a short study (median completion time < 7 minutes). People were told that they would read scientific summaries and make judgments about the authors of such texts.

### Participants and Power

Based on this study's preregistration (https://aspredicted.org/C3K_T31), 164 participants were required to detect a small effect powered at 80% in a within-subjects study ($f$ = 0.10, α = two-tailed, three measurements). A total of 274 participants were recruited to ensure enough participants were in the study. Most participants self-identified as men ($n$ = 139; 50.7%; women $n$ = 127, other $n$ = 7), they were 36.74 years old on average ($SD$ = 12.47 years), and were mostly White ($n$ = 190; 69.3%). On a 7-point political ideology scale (1 = extremely liberal, 7 = extremely conservative), participants leaned liberal ($M$ = 2.97, $SD$ = 1.63).

### Procedure

Five pairs of stimuli from Study 1b were selected for the experiment, having had the greatest difference in common words scores between the PNAS and GPT texts. Participants were randomly assigned to read stimuli from three out of a possible five pairs (see the online supplement for the stimuli texts), and within these randomly selected pairs, participants were randomly assigned to the GPT (simple) or PNAS (complex) version of each pair. Participants were told to read each summary of a scientific paper and then answer questions below each summary. They were specifically told "we are not expecting you to be an expert in the topic discussed below. Instead, make your judgments based on how the summary is written."

Finally, participants made various perceptions of the author (e.g., intelligence, trustworthiness) based on prior work (14, 15, 38), judgments about the identity of who wrote the scientific summary (AI or human), and assessed the complexity in each text as a manipulation check. The order of these measures was randomized, and items within each block were randomized as well. Experiments in this paper were approved by the author's university IRB.

### Measures

#### *Manipulation Checks*

Based on prior work (15, 38), three questions asked participants to rate how clear ("How clear was the writing in the summary you just read?"), complex ("How complex was the writing in the summary you just read?"), and how well they understood each scientific summary ("How much of this writing did you understand?"). Ratings for the first two questions were made on 7-point Likert-type scales from 1 = Not at all to 7 = Extremely. The third question ranged from 1 = Not at all to 7 = An enormous amount.

***Author Perceptions***

Participants made three ratings about the author of each scientific summary: (1) "How intelligent is the scientist who wrote this summary?", (2) "How credible is the scientist who wrote this summary?", and (3) "How trustworthy is the scientist who wrote this summary?" As a collection, these dimensions were highly reliable (Cronbach's α = 0.88) and therefore, they were averaged to create a general author perceptions index, while also being evaluated individually. All items were measured on 7-point Likert-type scales from 1 = Not at all to 7 = Extremely.

***Author Identity Perceptions***

Participants were asked for their agreement with two questions: (1) This summary was written by a human, and (2) This summary was written by Artificial Intelligence. All items were measured on 7-point Likert-type scales from 1 = Strongly disagree to 7 = Strongly agree.

***Demographics***

Basic demographic data were obtained from each participant, including their age, gender, ethnicity, and political ideology.

**Analytic Plan**

Since there were multiple observations per participant, linear mixed models with random intercepts for participant and stimulus were constructed (49, 50). Descriptive statistics for key measures are in the online supplement.

## Study 2: Results

Manipulation checks were successful. Participants perceived the simpler GPT significance statements as clearer ($B$ = 1.47, $SE$ = 0.09, $t$ = 16.70, $p$ < .001, $R^2m$ = .211, $R^2c$ = .502)[3], less complex ($B$ = -1.50, $SE$ = 0.08, $t$ = -19.28, $p$ < .001, $R^2m$ = .275, $R^2c$ = .498), and they reported understanding more information in such summaries than the complex PNAS versions ($B$ = 1.48, $SE$ = 0.08, $t$ = 18.74, $p$ < .001, $R^2m$ = .229, $R^2c$ = .584).

Crucially, as the top of Table 1 reveals, GPT significance statements were perceived more favorably than PNAS significance statements overall ($t$ = 2.44, $p$ = .015). Analyses at the item level told a more nuanced story, however. GPT significance statements were perceived as more credible ($t$ = 3.95, $p$ < .001) and more trustworthy than PNAS significance statements ($t$ = 4.63, $p$ < .001), but they were also perceived as less intelligent ($t$ = -2.57, $p$ = .010).

Ironically, participants agreed less with the idea that GPT significance statements were written by AI ($t$ = -4.30, $p$ < .001), and more with the idea that GPT significance statements were written by humans ($t$ = 5.42, $p$ < .001). Complexity is therefore perceived as more of an AI trait than a trait of humanness.

Together, people generally perceived the writers of scientific summaries more favorably if the text was simple compared to complex. The next step for this progression of studies is a replication and extension with more stimulus pairs, and crucially, an evaluation of participants' understanding of the science they read. If AI-generated summaries can facilitate more scientific understanding than human-generated summaries via simple writing, such language effects might serve as a lightweight and scalable intervention to improve scientific literacy and knowledge.

---

[3] $R^2m$ = variance explained by fixed effects alone; $R^2c$ = variance explained by fixed and random effects. All values were calculated using the MuMIn package in R (51).

## Study 3: Method

People in the US who did not participate in Study 2 were recruited from Prolific and paid $3.25 for their time. Participants were told they would read scientific summaries, make judgments about the authors of such texts, and answer comprehension questions about what they read.

### Participants and Power

Based on this study's preregistration (https://aspredicted.org/P9G_CFR), a minimum of 122 participants were required to detect a small effect powered at 80% in a within-subjects study ($f$ = 0.10, α = two-tailed, five measurements). A total of 250 participants were recruited, and most participants self-identified as women ($n$ = 149; 59.6%; men $n$ = 95, other $n$ = 6), they were 37.16 years old on average ($SD$ = 13.15 years), and were mostly White ($n$ = 170; 68.0%). On the 7-point political ideology scale, participants leaned liberal ($M$ = 3.16, $SD$ = 1.65).

### Procedure

This within-subjects study was identical to Study 2 except for two critical details. First, the stimulus set increased by four times relative to Study 2 ($n$ = 20 stimulus pairs) and were selected based on having had the greatest difference in common words scores between the PNAS and GPT texts. Participants also made judgments about a random selection of 5 stimuli in Study 3 (out of 20) instead of 3 stimuli (out of 5) that were randomly shown in Study 2.

Second, after participants read a randomly selected GPT or PNAS significance statement from 5 randomly selected pairs, they were presented with a multiple-choice question that tested the overall meaning of what they read. To construct a single question that could be answered by participants who read either the GPT (simple) or PNAS (complex) version of each text, one large language model chatbot (Gemini by Google) was instructed to read the significance statements together, create a question that could be answered by both texts, and provide the answer (see the supplement for the verbatim prompt). To ensure the answer to each question was correct and reliable, two large language models not used in the creation of the question nor in the creation of the significance statements (Claude 3 Opus by Anthropic and Llama 3 by MetaAI) answered the multiple-choice question after being given the significance statements (see the supplement for this prompt as well). The three large language models achieved perfect agreement on the correct answer (Krippendorff's α = 1.00). Answer choices to the multiple-choice questions during the experiment were randomized. All stimuli, including multiple choice questions and answers, are located on this paper's OSF page.

Following the multiple-choice question and a page break, participants were told to summarize the text they read ("In your own words, please summarize the main idea and findings of the study you just read about. Be as detailed as possible."). The free-response question was on a separate page from the text that participants read to avoid copying-and-pasting content. Consistent with a coding scheme from prior work (52), two independent large language models (GPT-4o and GPT-4) coded the texts for their accuracy and were blind to condition (see online supplement for coding instructions). The large language models achieved substantial agreement (Cohen's κ = 0.70, squared-weighted). Due to some level of disagreement, however, average ratings between the two large language model were used as the final free response score for each text. The multiple-choice and free-response questions were combined into a comprehension index by standardizing the values of each variable and adding their scores.

## Study 3: Results

### Perceptions Measures

Manipulation checks were successful. Participants perceived the simpler GPT significance statements as clearer ($B$ = 1.09, $SE$ = 0.07, $t$ = 15.33, $p$ < .001, $R^2m$ = .103, $R^2c$ = .539), less complex ($B$ = -1.10, $SE$ = 0.07, $t$ = -16.79, $p$ < .001, $R^2m$ = .129, $R^2c$ = .514), and they reported understanding more information in such summaries than the complex PNAS versions ($B$ = 0.96, $SE$ = 0.06, $t$ = 15.67, $p$ < .001, $R^2m$ = .089, $R^2c$ = .624).

The perceptions-based results were more mixed in Study 3 compared to Study 2 (see the bottom of Table 1). Simpler GPT texts were rated as less intelligent than more complex PNAS texts ($t$ = -3.34, $p$ < .001) and GPT texts were rated as marginally more credible than PNAS texts ($t$ = 1.65, $p$ = .099). However, the relationship between text type and trustworthiness was not statistically significant ($t$ = 1.46, $p$ = .146). Replicating Study 2, simpler GPT texts were more likely to be perceived as human ($t$ = 4.08, $p$ < .001) and less likely to be perceived as AI ($t$ = -2.53, $p$ = .012) than complex PNAS texts.

**Comprehension Measures**

For the comprehension index, participants displayed greater comprehension and understanding of the science when reading simple GPT texts compared complex PNAS texts ($t$ = 5.80, $p$ < .001). The most robust measure of this index was the free response question, where large language models rated participants who read simple GPT texts as having more accurate summaries than those who read complex PNAS texts ($t$ = 8.34, $p$ < .001).

Did comprehension differences also manifest in the participants' writing? To explore this question, participant summaries were analyzed with variables from LIWC: (1) adjectives, and (2) concreteness. Adjectives describe the level of detail in a text (46, 53) and concreteness considers the degree to which people are making direct, specific, and tangible references compared to indirect, broad, and abstract references (37, 38, 54, 55).[4]

Participants who read GPT texts were more detailed in their writing ($B$ = 0.86, $SE$ = 0.32, $t$ = 2.70, $p$ = .007, $R^2m$ = .005, $R^2c$ = .128) and had a more concrete writing style ($B$ = 0.24, $SE$ = 0.10, $t$ = 2.50, $p$ = .013, $R^2m$ = .005, $R^2c$ = .153) than participants who read PNAS texts. Taken together, not only did participants summarize the science better after reading simpler GPT texts compared to complex PNAS texts, but their writing was also more detailed and concrete, demonstrating additional downstream benefits of communicating science simply via generative AI.

## General Discussion

The current work explored the potential of generative AI to simplify scientific communication, enhance public perceptions of scientists, and increase the public's understanding of science. While lay summaries from a top general science journal, PNAS, were linguistically simpler than scientific summaries, the degree of difference between these texts could be enlarged and improved. Generative AI assisted in making scientific texts simpler and more approachable compared to the human-written versions of such summaries. Therefore, this paper is notable given current challenges of scientific literacy and the disconnect between scientific communities and the public — AI is indeed better at communicating like a human (or the intentions of writing simply) than humans (46, 56). As prior work suggests, decreasing trust in scientists and scientific institutions, exacerbated by complex communication barriers, call for inventive solutions that are scalable and relatively inexpensive. Those that are offered here, particularly through generative AI and simple writing, represent one potential pathway toward more approachable and improved science communication.

These data build on a body of fluency research and provide empirical support for the hypothesis that linguistic simplicity, facilitated by AI, can significantly influence public perceptions of scientists and crucially, also improve the public understanding of science. Generative AI, specifically large language models like GPT-4, can produce scientific summaries that are not only simpler, but also more accessible to lay audiences compared to those written by human experts. These results align with a broader scientific narrative (and interest) that advocates for clearer and more direct communication strategies in science dissemination (57).

---

[4] Consistent with prior work (37, 38, 54), concreteness was measured as an index by standardizing the rate of articles, prepositions, and quantifiers from LIWC, adding their values to create a single score. High scores indicate a person with more concrete thinking and writing style compared to low scores.

The implications of this paper are twofold. First, the results suggest that using AI in scientific communication can mend scientific communities and the general public. This could be particularly beneficial in a time where science is increasingly central to everyday decision-making but is also viewed with skepticism or deemed inaccessible by non-experts. Second, the increased readability and approachability of AI-generated texts might contribute to a higher engagement with and understanding of scientific content, thereby creating a more informed public. The experimental evidence supporting this contention is encouraging and deserves additional treatment in future research.

It is also important to underscore that the comprehension findings from Study 3 are timely for several reasons. The results provide support for the idea that linguistic simplicity can facilitate positive downstream perceptions and behaviors (18), especially for content that average readers likely find complex at the onset like science writing. The fact that AI-generated texts facilitated more detailed and concrete summaries of science than human-generated texts highlights how language simplicity may also enhance cognitive processing and people's retention of scientific information. If one goal of science is to create an informed and knowledgeable public, it is imperative to seriously consider tools that may help to improve the comprehension of scientific information like generative AI. Therefore, these results suggest using generative AI for science communication can democratize access to and the understanding of science. This is particularly crucial because scientific literacy is essential for key decision-making domains like in health, politics, and many others where generative AI are already impactful (58, 59). Altogether, by improving the clarity and approachability of scientific texts, AI has the potential to help the public engage with, appreciate, and understand science.

Despite the many positive outcomes and effects reported across studies, it is important to acknowledge that the simpler-is-better hypothesis (18) and simple writing heuristic (21) were not universally supported across measurements and studies. Regarding measurement, while AI-generated summaries were rated higher in terms of credibility and trustworthiness, they were also perceived as less intelligent (Study 2). This inconsistency, where all perceptions did not operate in the same direction, underscores the complex interplay between content simplicity and perceived expertise, suggesting that while simpler language can enhance understanding and trust, it might simultaneously reduce perceived intelligence. In science, people may be perceived as smart but untrustworthy and not credible, which suggests a one-size-fits-all model of the relationship between complexity and person-perceptions is perhaps inaccurate.

Regarding mixed results across experiments, Study 3 effects were consistent with Study 2 for intelligence, but only marginally significant for credibility and not statistically significant for trustworthiness. It is possible that with more sources of variation (e.g., four times as many stimuli in Study 3 compared to Study 2), content-related heterogeneity introduced a nontrivial source of variation that impacted average perceptions of scientists. That is, Study 3 may have revealed boundary conditions for the relationship between simplicity and perceptions of scientists, where the effects are more prominent in certain domains or content areas of science than others. It is also possible that with more variation across stimuli, the effect sizes are smaller than anticipated by the *a priori* power analysis. This is a reasonable explanation for the mixed results because Study 2 used 5 pairs of stimuli that had the greatest difference in common words, and Study 3 used 20 pairs with the greatest difference in common words. With less of a difference in common words as the number of pairs increased, effect sizes and simplicity's impact on perceptions might be attenuated. Therefore, future work should examine content and effect size heterogeneity in future work, yet it was still encouraging to see that the directions of the perceptions effects were still consistent across the two experiments.

**Long-Term Implications and Ethical Considerations**

Finally, it is important to elaborate on the possible long-term implications of using AI-generated summaries on public trust and engagement with science. If AI-generated summaries lead to better comprehension of scientific content, as demonstrated in Study 3, this could contribute to incremental improvements in scientific literacy among the public over time. The finding that simpler, AI-generated summaries led to higher perceptions of credibility and trustworthiness of the authors (Study 2) also suggests that presenting simple (versus complex) science may help address declining trust in scientists and scientific institutions (6–8). By making scientific content more approachable and easier to understand, AI-generated summaries might improve overall perceptions of science and how much people care about scientific information for pressing issues (e.g., climate change).

Despite these positive long-term implications, it is important to highlight several ethical considerations as well. One unintended consequence of simplifying science could be the loss of nuance or depth in the public's understanding of complex issues. In some cases, simply-written science might be heuristically-processed or encoded as science of less importance. For instance, if climate science is written in a simple manner, perhaps some people may think climate change is less pressing or less critical to understand, which could lead to skimming the research (22), confusion, or misinterpretations of climate issues. This might be especially problematic if people only read AI-generated summaries and not the research in full.

There are also ethical concerns about transparency and disclosure in scientific publishing with AI (60, 61). Should readers be informed that a scientific summary is AI-generated? Journals may be forced to lead in this regard if they decide to provide AI summaries of academic work. Relatedly, badges promoting open science practices (e.g., preregistration) have helped to improve perceptions of trust in science and scientists (9, 62), suggesting the transparent disclosure of AI-generated text might be helpful to mitigate reader concerns with such entities. Investigating how the public feels about AI-generated scientific summaries, with and without disclosure via a mechanism like badges, is an important area of future work.

Finally, AI-generated summaries are based on training data that could perpetuate or amplify biases in science or communication, in general. In just one example, consider a hypothetical large language model that was trained on medical papers and discussions of how people of different demographic backgrounds experience pain. Due to longstanding biases in medical research and practice, these training data likely contain examples of pain being dismissed or misdiagnosed for women and women of color (63–65). AI-generated summaries of the science on such issues might therefore reinforce biases where women and women of color are less likely to have their health concerns taken seriously. Together, there is a need to ensure AI systems, and the human tendencies that underly their development (66), are examined for the likelihood that they may exacerbate societal disparities.

**Limitations and Future Directions**

Future research should aim to explore how different domains of science (e.g., communicating about health, communicating about climate) might uniquely benefit from AI-mediated communication (67). Studies could investigate the long-term impact of AI-mediated communication strategies on public engagement with science and scientists. Texts from only one journal were used in this paper across studies and therefore, texts from other journals should be used as well to alleviate generalizability concerns. As a general science journal that publishes high-impact research, however, using PNAS for this paper was purposeful and helped to ensure fluency effects were investigated across core domains of scientific inquiry. Still, general science, flagship journals across various fields, and specialty journals might be used in future research to assess how well the patterns across studies hold. Future work may also look to use large language models other than those created by OpenAI (e.g., Llama by Meta, different open source models) to examine how these effects generalize.

Finally, while generative AI have demonstrated proficiency in creating scientific summaries that meet or exceed human expert abilities (34), future research might consider having experts rate AI-generated summaries along various scientific credibility dimensions to ensure the most critical parts of the work are being reviewed. This work will be instrumental in helping to ensure AI presents humans with not only accurate, but also relevant information that the authors themselves want to be represented in a summary of their work.

**Table 1**

*Estimated Marginal Means from Linear Mixed Models Across Experiments (Study 2 and Study 3)*

| | GPT (simple) | | PNAS (complex) | | | | | |
|---|---|---|---|---|---|---|---|---|
| Study 2 (*N* = 274 participants) | *M* | *SE* | *M* | *SE* | *t* | *p* | $R^2m$ | $R^2c$ |
| Perceptions index | 4.81 | 0.06 | 4.68 | 0.06 | 2.44 | .015 | 0.004 | 0.576 |
| Intelligence | 5.00 | 0.06 | 5.15 | 0.06 | -2.57 | .010 | 0.005 | 0.501 |
| Credibility | 4.72 | 0.07 | 4.47 | 0.07 | 3.95 | < .001 | 0.011 | 0.548 |
| Trustworthiness | 4.70 | 0.06 | 4.42 | 0.06 | 4.63 | < .001 | 0.015 | 0.558 |
| Perceived as human | 4.80 | 0.07 | 4.29 | 0.07 | 5.42 | < .001 | 0.033 | 0.165 |
| Perceived as AI | 3.68 | 0.07 | 4.10 | 0.07 | -4.30 | < .001 | 0.021 | 0.166 |
| Study 3 (*N* = 250 participants) | *M* | *SE* | *M* | *SE* | *t* | *p* | $R^2m$ | $R^2c$ |
| Perceptions index | 4.86 | 0.07 | 4.87 | 0.07 | -0.06 | .955 | 0.000 | 0.649 |
| Intelligence | 5.07 | 0.07 | 5.24 | 0.07 | -3.34 | < .001 | 0.004 | 0.585 |
| Credibility | 4.81 | 0.07 | 4.73 | 0.07 | 1.65 | .099 | 0.001 | 0.572 |
| Trustworthiness | 4.71 | 0.07 | 4.64 | 0.07 | 1.46 | .146 | 0.001 | 0.612 |
| Perceived as human | 4.58 | 0.07 | 4.27 | 0.07 | 4.08 | < .001 | 0.012 | 0.235 |
| Perceived as AI | 3.90 | 0.08 | 4.08 | 0.08 | -2.53 | .012 | 0.004 | 0.333 |
| Comprehension index | 0.19 | 0.13 | -0.22 | 0.13 | 5.80 | < .001 | 0.017 | 0.448 |
| Multiple choice | 0.61 | 0.26 | 0.40 | 0.26 | 1.59 | .112 | 0.002 | 0.319 |
| Average free-response coding | 0.92 | 0.04 | 0.72 | 0.04 | 8.34 | < .001 | 0.028 | 0.583 |

*Note*. Each model contains a random intercept for participant and stimulus. $R^2m$ = variance explained by fixed effects (condition) alone. $R^2c$ = variance explained by fixed effects and random effects. The model for the multiple-choice variable was a binary logistic mixed effects regression. In Study 2, participants were randomly assigned to 3 out of 5 stimulus pairs, and within each pair, randomly assigned to the GPT (simple) or PNAS (complex) version of text. In Study 3, participants were randomly assigned to 5 out of 20 stimulus pairs, and within each pair, randomly assigned to the GPT (simple) or PNAS (complex) version of text.

**Figure 1**

*Distributions of Key Comparisons in Study 1b*

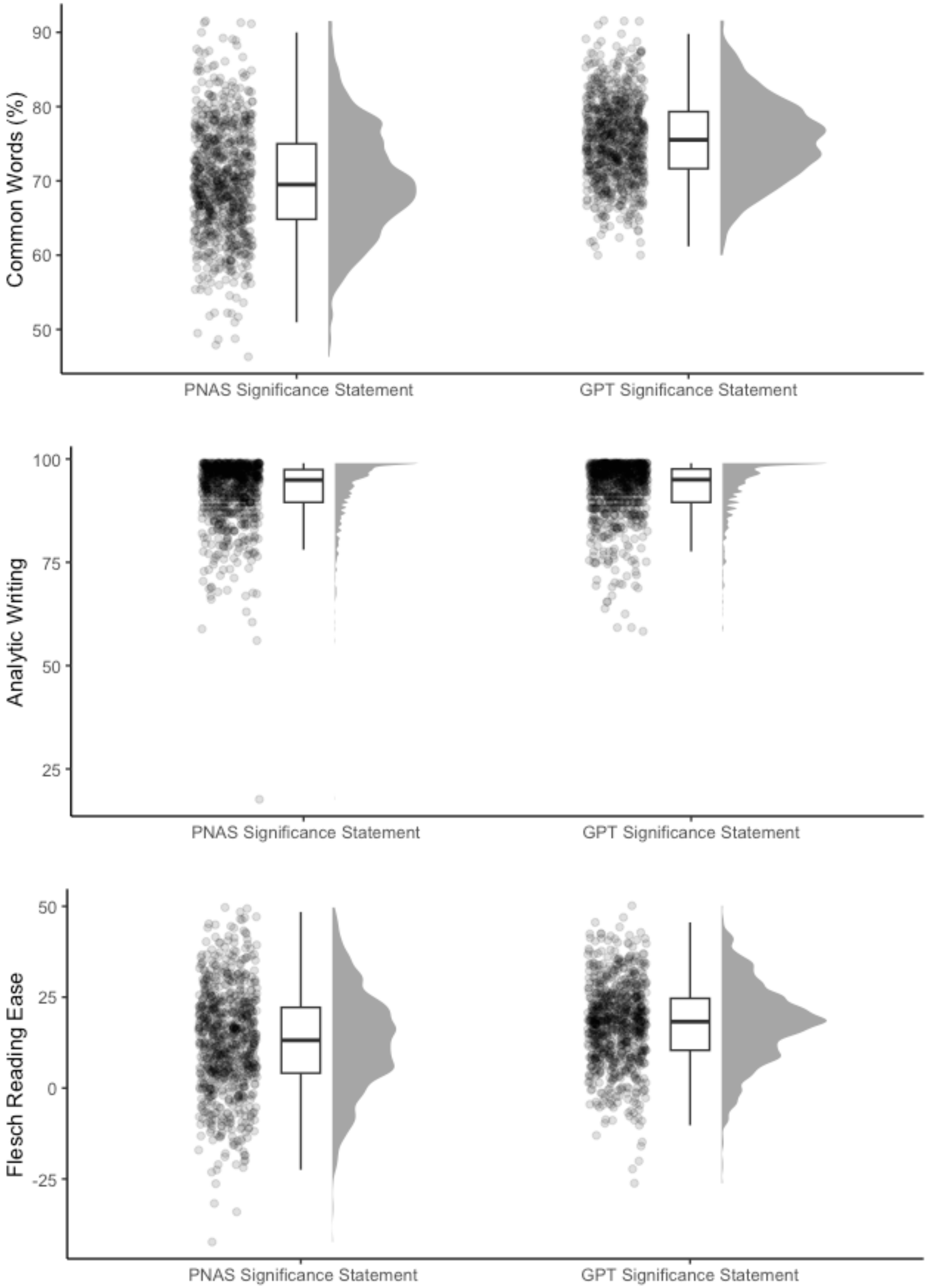